%% file: main.tex
\documentclass{article}
\usepackage[preprint]{neurips_2024}

\usepackage[utf8]{inputenc} 
\usepackage[T1]{fontenc}    
\usepackage{hyperref}       
\usepackage{url}            
\usepackage{booktabs}       
\usepackage{amsfonts}       
\usepackage{nicefrac}       
\usepackage{microtype}      
\usepackage{xcolor}         
\usepackage{graphicx}
\usepackage{algorithm}
\usepackage{algorithmicx}
\usepackage{amsmath}
\newcommand{\vx}{\mathbf{x}}

\newcommand{\lora}{LoRA}
\usepackage{authblk}
\usepackage{setspace} 

\newcommand{\emailfootnote}{\textsuperscript{*}}
\author[1]{Nirat Saini \emailfootnote}
\author[2]{Navaneeth Bodla}
\author[2]{Ashish Shrivastava}
\author[2]{Avinash Ravichandran\vspace{0.7em}}
\author[2]{Xiao Zhang}
\author[1]{Abhinav Shrivastava}
\author[2]{Bharat Singh}
\affil[1]{University of Maryland, College Park}
\affil[2]{Cruise LLC}

\title{InVi: Object Insertion In Videos Using Off-the-Shelf Diffusion Models}

\begin{document}

\maketitle
\footnotetext[1]{work done during internship at Cruise. Corresponding email: nirat@umd.edu.}
\input{abstract}

\input{introduction}

\input{related_works}

\begin{figure}[!t]
\centering{
\includegraphics[width=\linewidth]{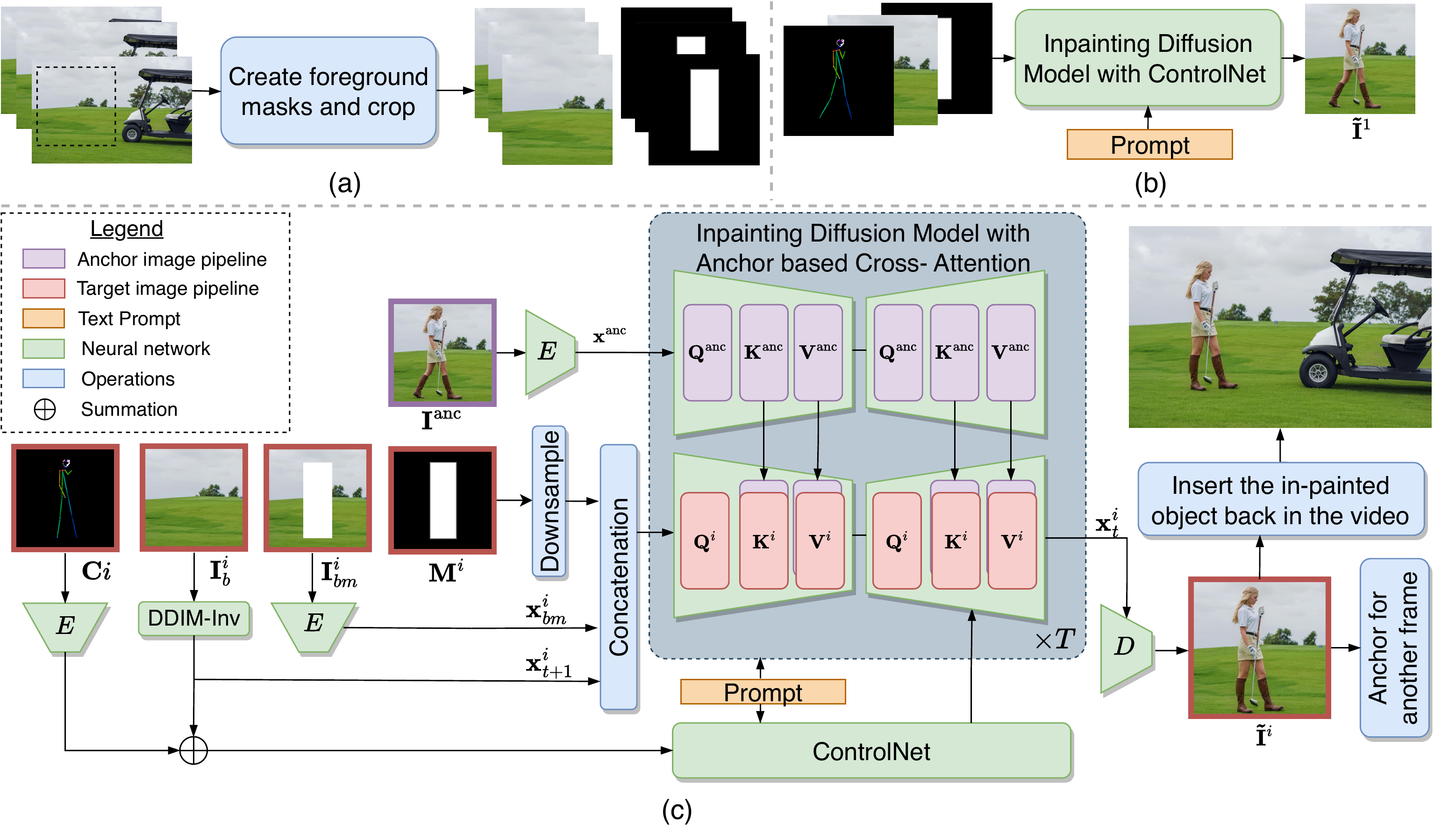}}
\vspace{-15pt}
\caption{
InVi inference pipeline: (a) Given a video and object bounding boxes, first, we crop a region around the bounding box which is inpainted. 
(b) Next, we use a ControlNet-based inpainting diffusion model to inpaint the cropped region in the first frame. (c) To ensure temporal consistency when inpainting subsequent frames, we employ the previous frame as an anchor image. 
This is achieved by adapting the self-attention block of the denoising U-Net with extended attention.
Specifically, we augment the Keys and Values of the current frame being inpainted with those of the anchor frame, allowing for consistent appearance. 
Finally, the inpainted crop is seamlessly integrated back into the input video. }
\label{fig::pipeline}
\end{figure} 

\input{method}

\input{experiments}

\input{conclusion}


\bibliographystyle{plainnat}
\bibliography{egbib}

\newpage
\appendix
\input{suppl}


\end{document}

%% file: abstract.tex
\vspace{-1em}
\begin{abstract}
\vspace{-1em}
We introduce InVi, an approach for inserting or replacing objects within videos (referred to as inpainting) using off-the-shelf, text-to-image latent diffusion models.  InVi targets controlled manipulation of objects and blending them seamlessly into a background video unlike existing video editing methods that focus on comprehensive re-styling or entire scene alterations.
To achieve this goal, we tackle two key challenges. 
Firstly, for high quality control and blending, we employ a two-step process involving inpainting and matching. 
This process begins with inserting the object into a single frame using a ControlNet-based inpainting diffusion model, and then generating subsequent frames conditioned on features from an inpainted frame as an anchor to minimize the domain gap between the background and the object. 
Secondly, to ensure temporal coherence, we replace the diffusion model's self-attention layers with extended-attention layers. The anchor frame features serve as the keys and values for these layers, enhancing consistency across frames.
Our approach removes the need for video-specific fine-tuning, presenting an efficient and adaptable solution. 
Experimental results demonstrate that InVi achieves realistic object insertion with consistent blending and coherence across frames, outperforming existing methods.
\end{abstract}

%% file: introduction.tex
\section{Introduction}
\label{sec:intro}
\vspace{-1em}
The emergence of image and video generation algorithms has opened up exciting new possibilities for utilizing generated data across various domains, including media production, AR/VR, and synthetic data for model training \cite{rombach2022high,guo2023animatediff,PNVR_2023_ICCV,ramesh2022hierarchical,esser2023structure, Shrivastava_2017_CVPR}. 
However, unconstrained text-to-image/video generation suffices only in a limited set of scenarios. 
In practice, there is often a need for enhanced control over image/video generation processes, encompassing aspects such as character consistency, pose, and beyond. 
This need has prompted the development of numerous algorithms in the image generation domain, including inpainting \cite{lugmayr2022repaint,rombach2022high}, \lora \cite{ruiz2023dreambooth,hu2022lora}, and ControlNet \cite{controlnet}. 
These techniques ensure that the generated images adhere to constraints such as background, style, and pose. 
In the realm of video generation, algorithms such as \cite{tokenflow2023,cao2023masactrl,wu2023tune} have addressed the demand for control, but many predominantly focus on comprehensive restyling of entire videos rather than the nuanced task of inserting or replacing specific objects within the video -- a process commonly known as inpainting. 
Furthermore, while some approaches tackle object manipulation, they often extend changes to the entire scene's background rather than solely concentrating on modifying the subject.
\begin{figure}[t]
\centering{
\includegraphics[width=\linewidth]{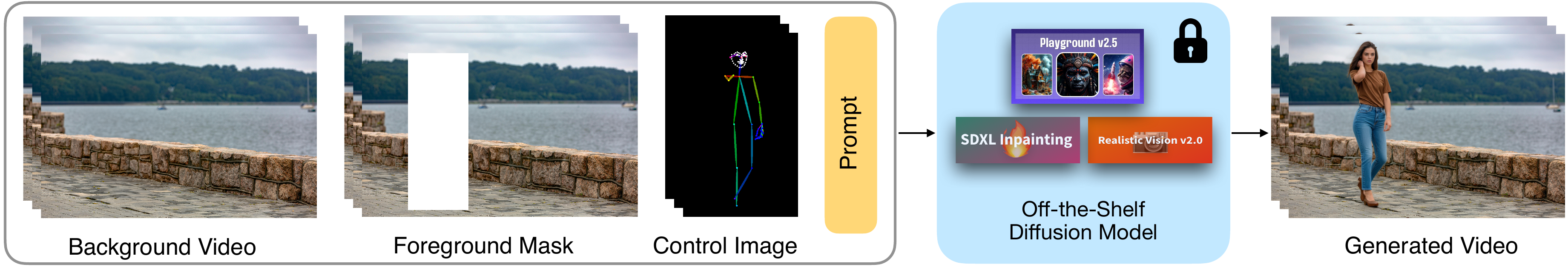}
\caption{InVi inserts objects into a background video using a foreground mask, a control signal (e.g., pose, canny, depth map), and a text prompt by leveraging off-the-shelf diffusion models. It ensures that the inserted object aligns semantically with the text, is temporally coherent in time, and also conforms spatially to the control signal.}
\label{fig::intro}
}
\end{figure} 
In this work, we focus on the tasks of adding and replacing objects in a video (Figure~\ref{fig::intro}). 
Unlike recent techniques such as those presented in \cite{tokenflow2023, wu2023tune}, we choose text-to-image diffusion models instead of text-to-video diffusion models, as the latter necessitate significant modifications for our specific task. 
Moreover, by building upon text-to-image models, we circumvent the requirement for training on extensive video datasets and can leverage a wide array of established text-to-image models spanning various domains, including anime, art, photography, autonomous driving, and more. 
This strategic choice enables us to take advantage of pre-trained conditional models like inpainting \cite{rombach2022high}, \lora \cite{ruiz2023dreambooth,hu2022lora}, ControlNet \cite{controlnet}, and seamlessly integrate them into our algorithm.

Existing approaches for video editing exhibit shortcomings, such as not generating all the frames \cite{tokenflow2023} or requiring expensive per-video fine-tuning \cite{wu2023tune}. 
Methods like Tokenflow \cite{tokenflow2023}, which opt for a joint synthesis approach, however, generates only a subset of the required frames and rely on optical flow to generate the remaining ones.
This limitation arises from the challenge of synthesizing all frames jointly, which becomes increasingly challenging due to GPU memory limitations, leading to performance degradation as the number of frames increases.
On the other hand, methods like Tune-a-Video \cite{wu2023tune} require additional temporal layers and fine-tuning on the target video, leading to significant latency.

To tackle these challenges, we introduce InVi, a novel method for inpainting objects in videos. 
Leveraging off-the-shelf text-to-image latent diffusion models, our approach seamlessly applies to videos of any duration, eliminating the requirement for individual fine-tuning for each video. 
In addressing object inpainting in videos, our method addresses two primary challenges:
(1) Ensuring realistic blending of the inserted object in the target video, avoiding a resemblance to its appearance in the source image. (2) Ensuring consistency across frames during the video synthesis process.

To achieve a seamless integration of the source image into the target image, InVi introduces a two-step inpaint and match process. 
Initially, the object is inserted into a single video frame, leveraging the effectiveness of image-based inpainting. 
Subsequently, the inpainted frame serves as the reference for generating subsequent frames, ensuring that video synthesis is conditioned on features within the domain of the target video rather than the source image alone. 
To maintain coherence across frames, InVi employs an auto-regressive architecture with extended-attention to incorporate features from the preceding frame while generating the current frame.  
Through experiments conducted on several videos from the DAVIS dataset and our own test set, which includes novel object insertion scenarios, we observe that InVi outperforms other methods by more than 40 points in background consistency metrics and is the preferred choice in nearly 70\% of the videos in our user study.

%% file: related_works.tex
\section{Related Works}
\textbf{Conditional video generation and editing}: 
Based on the progress in generating images from text with diffusion models \cite{saharia2022photorealistic,ramesh2022hierarchical,rombach2022high}, there has been an increase in works that address  video generation ~\cite{animatediff, controlavideo, wu2023tune}. This has facilitated the creation of videos from textual descriptions, which can be further refined to achieve video-to-video generation by using attributes derived from initial video inputs. For instance, Gen-1 \cite{esser2023structure} utilizes estimated depth as a conditioning factor, while VideoComposer \cite{wang2023videocomposer} uses a broader array of inputs such as depth, motion vectors, and sketches. However, most of these methods need explicit training on videos for learning motion~\cite{animatediff, controlavideo}, and ensuring that these models generalize to arbitrary motion patterns requires access to carefully curated large video datasets, which are relatively fewer (or non-existent) than those available for images \cite{schuhmann2022laion}. Additionally, substantial computational resources are required for the development of these models and their derivatives for conditional generation. To the best of our knowledge, there does not exist a text to video model which is trained end-to-end, which supports inpainting objects in videos, while providing support for using auxiliary conditions like pose, depth, edgemaps, etc., as is commonly available for images. To overcome the challenges associated with training such complex models on videos, some approaches resort to single-image editing, subsequently extending these modifications across video sequences by identifying and applying edits to corresponding pixels throughout the frames and their efficacy hinges on robust tracking. Various methods \cite{yang2023rerender,vid_swap} have employed techniques such as optical flow, keypoint tracking, or other forms of motion detection to address this challenge. However, these techniques are hard to scale to long videos, for consistent appearance changes in objects.

\noindent \textbf{Adapting Image Models for Video to Video tasks:} Many methods have extended image-to-image models for swapping objects in videos. 
For instance, \cite{khachatryan2023text2video} modifies self-attention mechanisms in diffusion models, while \cite{wu2023tune} conducts per-video fine-tuning and employs inversion-denoising techniques for editing purposes. MasaCtrl \cite{cao2023masactrl}, originally developed for image editing tasks, has been extended to video generation tasks and leverages the first frame generated as a reference to synthesize subsequent frames in the video sequence. 

\cite{liu2023video} and Fate-Zero \cite{fatezero} adapt image-to-image pipelines \cite{hertz2022prompt, tumanyan2023plug, brooks2023instructpix2pix} for video editing by introducing modifications to cross-frame attention modules, incorporating null-text inversion, and more. However, most existing methods are limited to generating very short video clips. TokenFlow \cite{tokenflow2023} produces keyframes and employs a nearest-neighbor field on diffusion features to extend keyframe attributes to remaining frames. However, as the video length increases, interpolation performance may degrade due to accumulated interpolation errors over time. In contrast, our model enhances spatio-temporal attention \cite{khachatryan2023text2video,liu2023video,fatezero} with anchor-based cross-frame attention, enabling the generation of long videos with any desired number of frames. Our work also differs from TokenFlow \cite{tokenflow2023} in its support for inpainting. \cite{tokenflow2023} does not support in-painting, as it is tailored to preserve the structure and motion of the original video and cannot handle edits like changing the size, shape, pose or motion patterns of objects. We use similar ideas of latent inversion of the source video, but they can be of a video from a different domain, and we can use it's pose or canny features to inpaint a similar object in new videos. This ensures sharp and consistent object insertion in new videos, while \cite{tokenflow2023} fails in maintaining sharpness of a new object, due to its optical flow propagation in the latent space.

%% file: method.tex
\section{InVi}
We build upon the concepts of Latent Diffusion Models~\cite{ho2020denoising, rombach2022high}, DDIM inversion~\cite{rombach2022high, tokenflow2023} and Lora~\cite{hu2022lora}. Readers are encouraged to refer to the methods or appendix for a more in-depth details.
Given an input video $\boldsymbol{\mathcal{I}} = [\mathbf{I}^1, ..., \mathbf{I}^n]$ comprising $n$ frames,  a text prompt $\boldsymbol{\mathcal{P}}$ describing the desired edit and a control sequence $\boldsymbol{\mathcal{C}} = [\mathbf{C}^1, ..., \mathbf{C}^n]$, InVi generates an edited video $\tilde {\boldsymbol{\mathcal I}} = [\tilde {\mathbf I}^1, ..., \tilde {\mathbf I}^n]$. 
As in  LDM~\cite{rombach2022high}, the video frames are converted to latent feature using an encoder, $E$, and the corresponding encoded features are denoted by $[\mathbf x^1, \dots, \mathbf x^n]$.
Similarly, the encoded features of the edited video are denoted by $[{\tilde{\mathbf  x}}^1, \dots, {\tilde{\mathbf  x}}^n]$.
The edited video aligns spatially with the control sequence $\boldsymbol{\mathcal{C}}$ and conforms to the semantic constraints outlined in $\boldsymbol{\mathcal{P}}$.
The text prompt, $\boldsymbol{\mathcal{P}}$, offers generic semantic guidance, influencing factors such as object appearance. Alternatively, the desired edit's appearance can be specified directly as an image instead of the text prompt, for which, we leverage LoRA~\cite{hu2022lora}. 
In contrast, the control sequence $\boldsymbol{\mathcal{C}}$ provides more nuanced control, such as pose or object shape.
Various methods exist for providing spatial control, denoted by $\boldsymbol{\mathcal{C}}$, such as depth maps, edge maps, and normal maps for generic objects, or human poses if the object is a person~\cite{controlnet}. 
Next, we will describe each of the steps in our pipeline in more detail.

\subsection{Generating the first-frame and pre-processing}
First, given the object's location in each frame via  bounding boxes, we extract a region of fixed resolution by expanding these bounding boxes, as illustrated in Figure~\ref{fig::pipeline}(a).
We then insert an object into the first frame, for which we rely on a ControlNet-based inpainting diffusion model.
This can be any off-the-shelf text to image inpainting model for prompt-based editing and personalized   model using \lora \cite{ruiz2023dreambooth} for  reference image based editing.  
Once one frame is edited using image inpainting pipeline, we use this generated image as an ``anchor'' denoted as $\mathbf{I^\text{anc}}$ and edit the remaining frames. 

To prepare the inputs for generating subsequent frames, we first pass the masked image through a VAE encoder, as done in prior work \cite{rombach2022high}, compressing it into a lower-dimensional input ($64\times64\times4$ in our experiments). 
This input is then concatenated with a suitably downsampled mask of identical dimensions ($64\times64$), indicating the area to be inpainted. 
In contrast to the inpainting pipeline in \cite{rombach2022high}, which combines these inputs with Gaussian noise (sized $64\times64\times4$) during inpainting, we utilize the output after DDIM inversion on background frames as input for the inpainting model. 
This step is crucial for maintaining video consistency, as DDIM inversion on the background frame ensures a consistent noise pattern across frames. 
Additional conditions such as pose or depth-map are provided to ControlNet, as outlined in \cite{controlnet}. A comprehensive wire diagram detailing all inputs for our pipeline is illustrated in Figure~\ref{fig::pipeline}(c).

\subsection{Temporally Consistent Frame Inpainting}
To propagate information from the edited anchor frame $\mathbf{I}^\text{anc}$ to another video frame, we propose to use cross-frame attention mechanisms, circumventing conventional methods such as optical flow or explicit point tracking. Given an anchor frame, $\mathbf{I}^\text{anc}$, we incorporate it as an additional input to the diffusion model and replace the self-attention mechanism in the model with cross-frame attention. 

Specifically, we use the anchor frame features to augment keys, denoted by $\mathbf K$, and values, denoted by $\mathbf V$, within the attention layers of the diffusion model. We denote the key and value matrices of the $i^{th}$ frame as $\mathbf K_{i, l, t}$ and $\mathbf V_{i, l, t}$, respectively, where $l$ is the layer index of the diffusion model and $t$ is the diffusion step. 
Similarly, we denote the key and value matrices of the model obtained when the anchor frame is passed to the model as  $\mathbf K^\text{anc}_{l, t}$ and $\mathbf V^\text{anc}_{l, t}$\footnote{For brevity, we will omit the subscripts $l$ and $t$ where context makes it clear.}, respectively. To edit $i$-th frame $\mathbf I^i$, we modify  the self-attention module to a cross-frame attention using the key and value vectors of anchor frames as follows:

\begin{equation*}
    l^{\text{th}} \text{layer feature}  = \texttt{Softmax} \left( \frac{\mathbf Q_{i, l, t} [\mathbf K_{i, l, t}, \mathbf K^\text{anc}_{l, t}]^T}{\sqrt{d}} \right) [\mathbf V_{i, l, t},\mathbf V^\text{anc}_{l, t}],  \; \forall l, \forall t \in [1, \dots, T].
\end{equation*}

Note that this augmentation does not change the network architecture and does not require any learning of new parameters. Our method, as shown in Figure \ref{fig::pipeline}(c), utilizes softmax-generated attention scores to integrate $\mathbf V^\text{anc}$ features from the anchor frame. 
This process effectively enforces the temporal correspondence between the current frame and the anchor frame, and facilitates the propagation of value features from the anchor frames to the current frame through the multiplication of attention scores with $\mathbf V^\text{anc}$.
By substituting the self-attention module with an anchor-based cross-frame attention mechanism, we achieve temporal consistency across the edited video frames.

  We could use one anchor frame for the entire video, however, this is not ideal as the background appearance and the pose of an object gradually evolves over time. Therefore, once we generate a frame $i$, it serves as the anchor for generating the next frame $i+1$. 
 This sequential process is described in Algorithm \ref{alg:video_generation}. 
\begin{algorithm}[H]
\small
    \caption{InVi: Object Insertion in Videos}
    \label{alg:video_generation}
    \textbf{Input:}
    \begin{algorithmic}
        \State ${\mathbf X}=[\mathbf x^1_b,\dots ,\mathbf x^n_b]$ \hfill $\triangleright$ Background video in latent space
        \State $\boldsymbol{\mathcal M}=[\mathbf M^\text{1},\dots ,\mathbf M^n]$ \hfill $\triangleright$ Downsampled input mask
        \State $\mathbf X_{bm}=[\mathbf x^1_{bm},\dots ,\mathbf x^n_{bm}]$ \hfill $\triangleright$ Masked background in latent space
        \State $\boldsymbol{\mathcal{C}}=[\mathbf C^1,\dots ,\mathbf C^n]$ \hfill $\triangleright$ Conditional inputs
        \State $\boldsymbol{\mathcal{P}}$, $\boldsymbol{\phi}$ \hfill $\triangleright$ Target text prompt, ControlNet-based inpainting model        
    \end{algorithmic}
    
    $\{ \mathbf{x}^i_t \}_{t=1}^T \gets \text{DDIM-Inv}[\mathbf x^i_b] \quad \forall i\in [1, \dots, n], \; \forall t \in [1, \dots, T]$ \\
    \textbf{For} $t=T,\dots,1$  \textbf{do}  
    \begin{algorithmic}
        \State $\tilde{\mathbf x}^1_t = \boldsymbol{\phi}(\mathbf x^1_t, \mathbf x^1_{bm}, \mathbf M^1, \mathbf C^1)$  
        \State $\mathbf K^\text{anc}_{l,t}, \mathbf V^\text{anc}_{l,t} \gets  \mathbf K_{1, l,t}, \mathbf V_{1, l,t} \qquad \forall l $ \hfill $\triangleright$ save first frame features in a cache
    \end{algorithmic} 
    
    \textbf{For} $i=2,\dots,n$ \textbf{do}
    \begin{algorithmic}
        \State
        \textbf{For} $t=T,\dots,1$  \textbf{do}
        \begin{algorithmic}
            \State load $\mathbf K^\text{anc}_{l,t}, \mathbf V^\text{anc}_{l,t} $ from cache 
            \State $\tilde {\mathbf x}^{i}_t \gets \boldsymbol{\phi}({\mathbf x^i_t, \mathbf x^i_{bm_t}, \mathbf M^i_t, \mathbf C^i_t, \mathbf K^\text{anc}_t, V^\text{anc}_t}) $ \hfill $\triangleright$ inpaint $i$-th frame with anchor features 
            \State save $\mathbf K_{i, l,t}, \mathbf V_{i, l,t}$
        \end{algorithmic}
$\mathbf K^\text{anc}_{l,t}, \mathbf V^\text{anc}_{l,t} \gets \mathbf K_{i, l,t}, \mathbf V_{i, l,t}$ \hfill $\triangleright$ Update cache with $i$-th frame features
    \end{algorithmic}
    
    \textbf{Output:} $\tilde{\mathbf X} =[\tilde{ \mathbf x}_1^1, \dots, \tilde{ \mathbf x}_1^n]$ \hfill $\triangleright$ \text{Latents for inpainted frames at $t=1$}
\end{algorithm}  

\subsection{Post-processing}
After inpainting  the object within the Region of Interest (RoI), an occasional subtle halo effect emerges, resembling a flickering square, in the vicinity of the inserted object. In the case of high-resolution videos, due to the limited training of base diffusion models on such resolutions (and an order of magnitude higher inference time), object inpainting can only be performed within a small RoI. The subtle differences which result from VAE based reconstruction are not very prominent (although noticeable) when the inpainted RoI is composed with the original frame but this gets amplified in a video as the object moves.
Consequently, to achieve seamless and efficient blending for high resolution videos, we adopt a multi-step approach. Initially, we extract the mask of the inserted object using grounding-DINO~\cite{liu2023grounding} (for detecting arbitrary classes) and SAM~\cite{kirillov2023segment} (getting object masks inside bounding boxes). Once the mask is obtained, we employ dilation to expand its boundary. Subsequently, we utilize Lama~\cite{suvorov2022resolution} to inpaint the pixels within this boundary, ensuring smooth blending throughout the video sequence as shown in Figure~\ref{fig:blending}. This comprehensive strategy enhances visual coherence and minimizes any artifacts or discrepancies resulting from the object insertion process. Note that for low resolution videos where the entire frame can be inpainted, we do not require this step.

%% file: experiments.tex
\begin{figure}[t]
\centering{
\includegraphics[width=\linewidth]{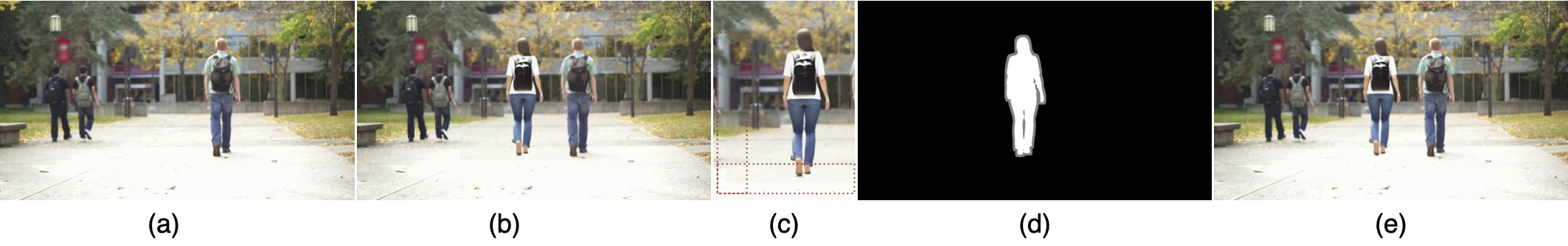}}
\vspace{-15pt}
\caption{Post-processing to remove flickering square artifacts. a) Background image. b) Initial image generated from our pipeline. c) Zoomed-in view revealing artifacts around the inserted object. d) A trimap is generated to facilitate seamless blending of the object into the background. e) Post-processed frame showcasing the final result after blending the inserted object with the background.}
\label{fig:blending}
\vspace{-15pt}
\end{figure} 
\section{Experiments}
Our method is evaluated across diverse datasets, including videos from the DAVIS dataset which are used in prior work~\cite{davis}, a selection of videos from the VIRAT surveillance dataset~\cite{VIRAT}, as well as human-centric videos sourced from YouTube. 
Additionally, we curate our own video footage featuring cars, traffic cones, falling balls, and various moving objects to further assess the robustness and efficacy of our method. 
To replicate synthetic assets suitable for insertion into a 3D scene using simulation engines for applications like surveillance, AR/VR, and autonomous driving, we adopt two approaches. 
Firstly, we gather videos with a static camera over extended durations, enabling the extraction of conditional inputs from earlier time frames. Alternatively, we employ object removal software (RunwayML) to artificially remove objects from scenes, allowing us to utilize conditional inputs from the original video. We attach examples of these in the supplementary material.
The spatial resolution of our videos (after cropping) is $384\!\times\!672$ or $512\!\times\!512$ pixels, and they consist of anywhere from $24$ to $200$ frames. Our evaluation dataset comprises of $30$ text-video pairs. 
When training a \lora-Dreambooth model, we train for $1200$ iterations with a rank of $96$, using a single reference image, without setting any regularization. In our experiments, we use the inpainting version of RealisticVision 5.0, which is based on Stable Diffusion 1.5 \cite{rombach2022high}. Our computational overhead (apart from DDIM inversion) is minimal compared to the per-frame baseline as we only double the FLOPs, and memory in the self-attention blocks of the transformer layers, while everything else remains the same.

\subsection{Baselines} 
We benchmark InVi against several video editing methods that swap objects while preserving their structure. These include: (1) Fate-Zero~\cite{fatezero}, a zero-shot text-based video editing method; (2) Tune-a-Video~\cite{wu2023tune}, which fine-tunes the text-to-image model on the given test video; and (3) TokenFlow~\cite{tokenflow2023}, which edits selected anchor frames and propagates the implicit flow from the keyframes to the rest of the video using an off-the-shelf propagation method. We employ PnP-Diffusion~\cite{pnp} based editing with TokenFlow.
These methods alter the entire frame and do not preserve the background. Since there are no existing video inpainting methods that utilize off-the-shelf diffusion models, we include two additional baselines to evaluate the inpainting performance: (1) Per-frame diffusion-based image inpainting baseline using ControlNet; and (2) a ControlNet~\cite{controlnet} based inpainting pipeline for TokenFlow.

\begin{table*}[t]
\caption{Quantitative Results for object swapping (on the left) and object insertion (on the right). Evaluation for background consistency, temporal appearance consistency, and alignment with prompts.}
\begin{minipage}[t]{.53\textwidth}
  \centering
  \small
  \renewcommand{\tabcolsep}{5pt}
  \renewcommand{\arraystretch}{1.2}
 \resizebox{\textwidth}{!}{
 \begin{tabular}{@{}lcccc@{}}
  \toprule
 & FateZero & Tune-a-Video & TokenFlow & InVi \\
\midrule
CLIP-Text & 0.30 & 0.31 & 0.32 & 0.33 \\
CLIP-Temp & 0.95 & 0.95 & 0.96 & 0.97\\
Back-L1 & 35.66 & 100.98  & 42.26 & 6.40 \\
\bottomrule
\end{tabular}}
 \end{minipage}
 \hfill
 \begin{minipage}[t]{.4\textwidth}
  \centering
  \small
  \renewcommand{\tabcolsep}{5pt}
\renewcommand{\arraystretch}{1.33}
 \resizebox{\textwidth}{!}{
  \begin{tabular}{@{}lccc@{}}
  \toprule
 & Frm+Inp & TokenFlow+Inp  & InVi \\
\midrule
CLIP-Text & 0.24 & 0.26 & 0.28 \\
CLIP-Temp & 0.96 & 0.97 & 0.98 \\
LPIPS & 0.07 & 0.05 & 0.02 \\
\bottomrule
\end{tabular}}
\end{minipage}
\vspace{-0.1in}
\label{tab:Quant}
\end{table*}
\begin{figure}[t]
\centering{
\includegraphics[width=\linewidth]{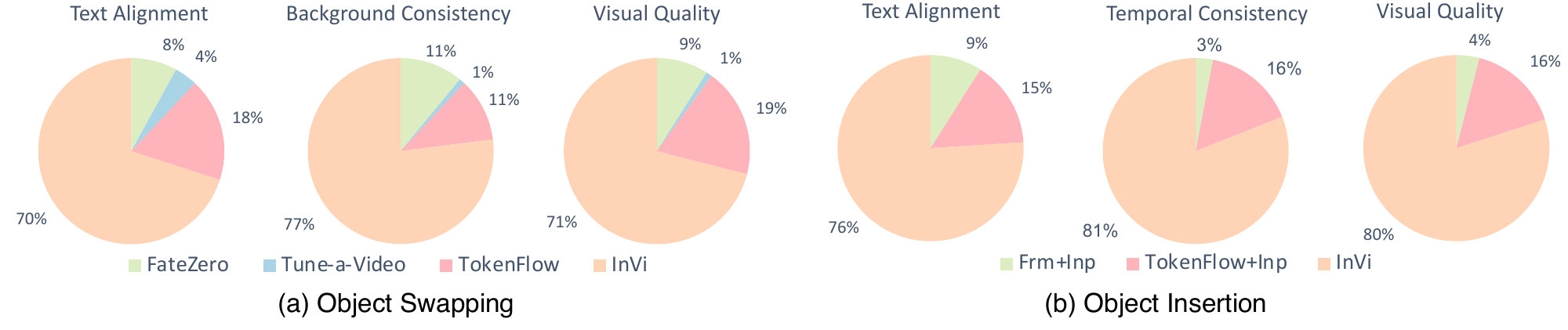}
\caption{User Preference Study: InVi Outperforms Baseline Methods in text alignment, background and temporal appearance consistency and overall video quality.}
\label{fig:user_study}}
\vspace{-0.2in}
\end{figure} 
\begin{figure}[t]
\centering{
\includegraphics[width=0.9\linewidth]{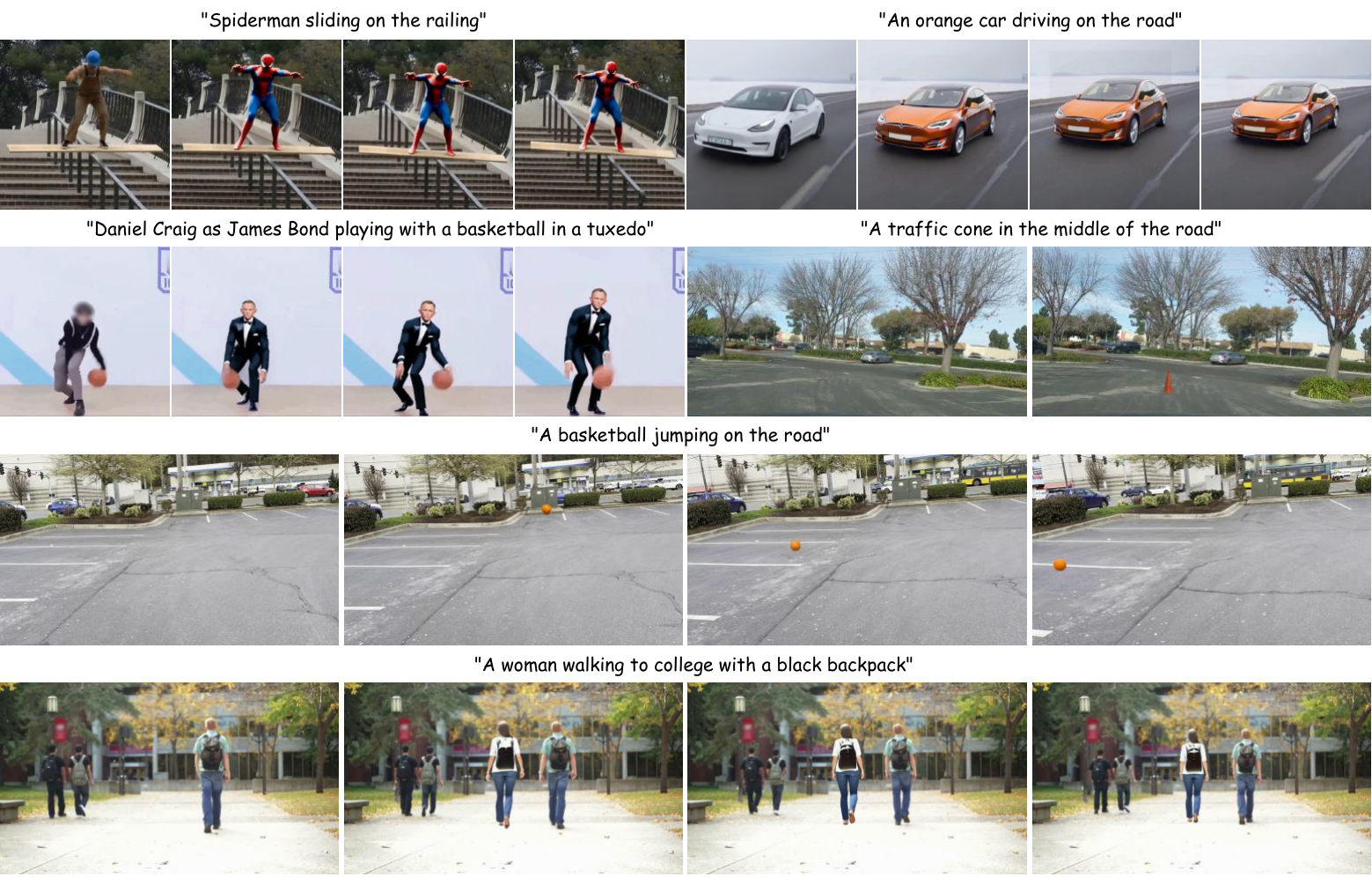}}
\caption{Qualitative results. The first image is a background frame from the video undergoing inpainting. Subsequent frames depict the video with the inserted object.}
\label{fig:results}
\vspace{-15pt}
\end{figure} 
\subsection{Quantitative Evaluation}
Following previous work\cite{tokenflow2023, fatezero}, we use several metrics to evaluate various aspects of our object editing and inpainting techniques. Firstly, we compute \textbf{CLIP-Text}, which represents the average CLIP feature similarity between the generated frames and the target prompt, serving as an indicator of video-text alignment. For assessing temporal consistency within the video, we utilize \textbf{CLIP-Temp}, which measures the similarity of consecutive frames and averages the results across the generated video.
Given the importance of maintaining background consistency while editing a specific object in videos, we use a background mask to evaluate \textbf{Back-L1} which is the average L1 distance between each pixel across corresponding frames of original video and edited video. Video editing is more common task, hence we compare with existing baselines which operate off-the-shelf without any training, for a fair comparison with our method. For inserting new objects in a video, all baselines and our method inpaint only the object, the background remains consistent. Hence, instead of \textbf{Back-L1} we use average \textbf{LPIPS}~\cite{lpips}, which is patch based perceptual similarity score across  consecutive frames of the video. Lower LPIPS means more similarity across frames.  Finally, in addition to objective metrics, we conduct a user study to gauge the alignment of the edited video quality with human preferences, covering aspects such as text alignment, background changes, temporal consistency, and overall impressions. 

\subsection{User Study} Video editing and inpainting is a subjective task, where quality of results cannot be evaluated with quantitative metrics alone. Hence, we also conduct a user preference study (with 15 users, 195 question responses), where users are shown videos of baselines and our method, and are asked to pick the video with best text alignment, background consistency (for edited video), temporal consistency (if the inpainted object is consistent in appearance across frames) and overall visual quality (least blurriness and extra artifacts). Figure~\ref{fig:user_study} shows that users prefer InVi across all  questions $\sim75\%$ times. While Tokenflow~\cite{tokenflow2023} is preferred $\sim$15\% of times across all the qualitative categories. More details can be found in supplemental materials.

\subsection{Qualitative Evaluation}
\begin{figure}[h]
\centering{
\includegraphics[width=0.87\linewidth]{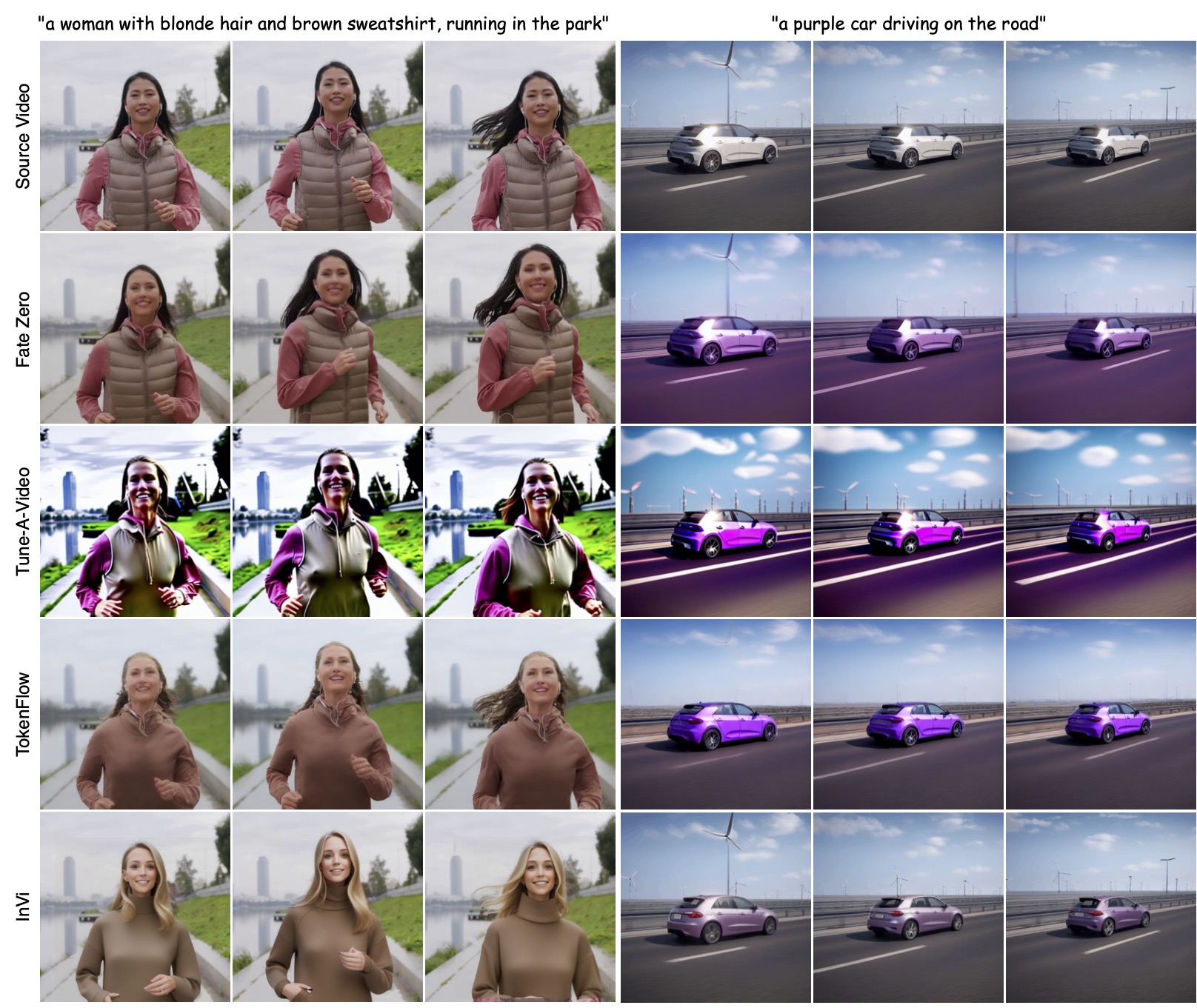}}
\caption{In our qualitative comparison, we contrast the performance of InVi with three baseline methods: FateZero, TokenFlow, and Tune-a-video. 
FateZero frequently diverges from the editing prompt, as seen in the woman running example. Meanwhile, both TokenFlow and Tune-a-video unintentionally modify the background. InVi, however, consistently yields results that closely align with the editing prompt while preserving the background.}
\label{fig:baseline}
\vspace{-15pt}
\end{figure}

As depicted in Figure~\ref{fig:baseline}, we conduct a comparative analysis of InVi against prominent baselines. Our approach, represented in the bottom row, demonstrates superior performance by closely adhering to editing instructions and ensuring temporal coherence in the edited videos. Conversely, other techniques often struggle to achieve both objectives simultaneously. Tune-A-Video~\cite{wu2023tune} expands a 2D image model into a video model and fine-tunes it to follow the video's movement closely. While effective for short clips, it encounters challenges in accurately capturing movement in longer videos, resulting in visual artifacts such as cartoonish appearances in the edited videos, as observed in the car example. Similarly, fate-zero also exhibits artifacts and deviates from the editing text-prompt closely. Although TokenFlow~\cite{tokenflow2023} yields reasonable results overall, it fails to perform well for inpainting. While it effectively edits rigid objects using flow, it struggles with inserting articulated moving objects like walking people. Moreover, all baselines exhibit inconsistencies in maintaining the background consistent with the source video, often modifying the background along with the object to be edited. Through a comprehensive user study and qualitative assessments, as shown in Table~\ref{tab:Quant} and Figure~\ref{fig:results}, we demonstrate that InVi excels in preserving background consistency while inserting new objects into the scene.

\begin{figure} [h]
\centering{
\includegraphics[width=0.87\linewidth]{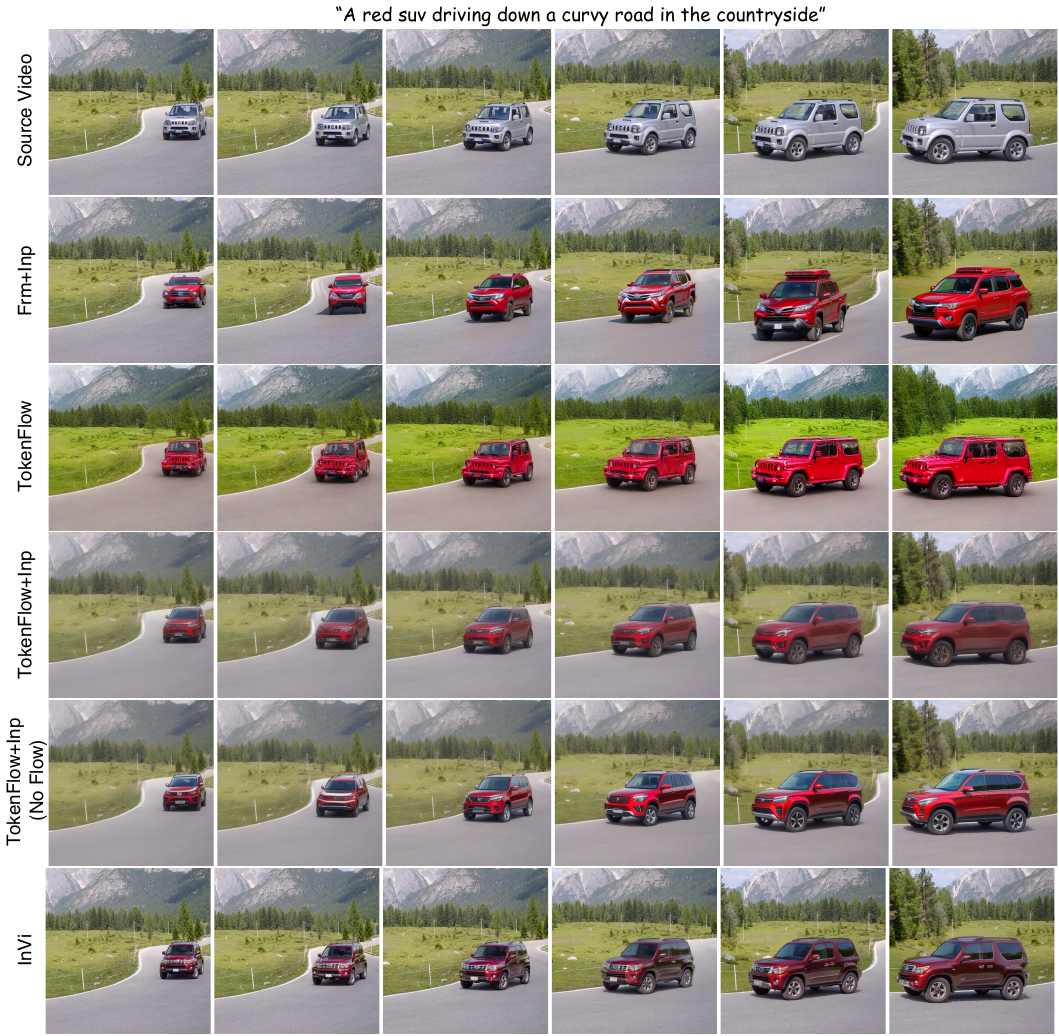}}
\caption{\textbf{Ablation experiments:}  We make simple changes to the baseline methods.  
Frm+Inp conducts frame-wise inpainting using a constant seed and prompt. Tokenflow preserves the exact structure of the original jeep (like preserves the grills and mostly changes the color). TokenFlow+Inp combines ControlNet along with an inpainting method, serving as a baseline for inpainting, but leads to blurry results. TokenFlow+Inp (No Flow) removes the nearest-neighbor field computation from Tokenflow, and keeps the sliding window based inpainting of 2 frames at a time. Finally, InVi, which surpasses these methods in terms of clarity, consistency, and sharpness, establishing itself as the preferred choice for inpainting tasks.}
\label{fig:abl1}
\vspace{-15pt}
\end{figure}
\subsection{Ablation}
\subsubsection{Advancing Beyond TokenFlow for Inpainting Tasks}
TokenFlow~\cite{tokenflow2023} primarily works for video editing tasks, and relies on optical flow in the latent space. However, as seen in Figure~\ref{fig:abl1} (Row 3), it results in color leakage in edited objects and unwanted color saturation changes in background colors. To compare with an inpainting pipeline, we modified TokenFlow to include a 9-channel inpainting based UNet for inference alongside ControlNet~\cite{controlnet}. This mitigates the color leakage issues and enhances background consistency, as seen in Row 4 of Figure~\ref{fig:abl1}, but leads to blurry and unrealistic video outputs. TokenFlow relies on two main components: (i) Extended-attention, which selects and edits 5-6 frames sparsely from the video, ensuring consistent appearance across all frames, and (ii) Flow propagation across other frames, which computes latents for unedited frames through interpolation in the latent space based on edited frames. Our hypothesis suggests that the blurriness observed in TokenFlow results from flow computation. We experiment with using only extended-attention with a sliding window, which results in sharper inpainted objects (Row 5). 

InVi edits one frame and recursively use the generated frames for editing the remaining video, which ensures consistency in appearance throughout the video. Because of the recursive approach, we do not need to jointly generate $K$ frames sampled across the video, but we only use the previously generated frame while generating the next frame. Hence, our memory usage only increases by a factor of $2$ compared to Tokenflow~\cite{tokenflow2023}.

%% file: conclusion.tex
\vspace{-1em}
\section{Conclusion and Future Work}
\vspace{-1em}
We presented a new approach to use text-to-image models for video inpainting tasks, using off-the-shelf models which operates without the need for video-specific training. By harnessing DDIM inverted latents extracted from the source video and incorporating the structural information of new objects via conditional ControlNet inputs, InVi seamlessly inpaints new objects into scenes. Utilizing anchor-frame based extended-attention for editing frames, InVi ensures both consistency in appearance and structure of the inserted object. Our method surpasses existing baselines, showcasing significant enhancements in temporal consistency and visual fidelity. Moreover, unlike prior methods, InVi efficiently handles longer videos with limited GPU memory and enables the insertion of dynamic objects without requiring an explicit motion module. One limitation of our work is that our method relies on 2D bounding boxes in each frame, which can either be provided by the user or estimated using the geometry of the scene. In future work, we plan to automate the generation of these boxes using GPT based layout generation techniques, so that it can be more broadly applicable. As our work builds upon existing image generation methods, we inherit both the positive and negative societal impact of such methods.

%% file: suppl.tex
\input{prelims}
\begin{figure}[tbh!]
\centering{
\includegraphics[width=0.85\linewidth]{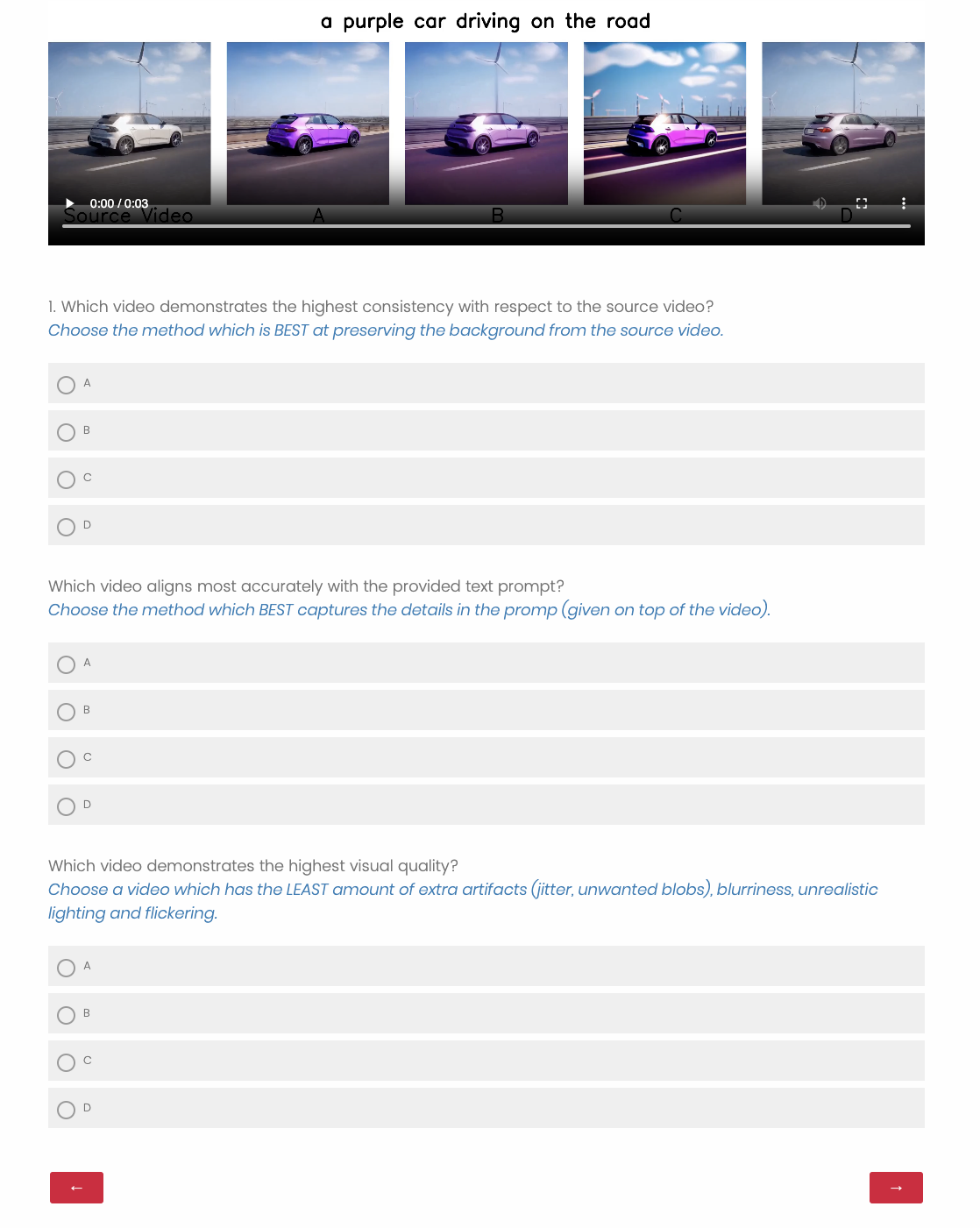}
\caption{Survey Preview for Object swapping videos. The users are shown 4 videos along with source video and prompt used for editing, to answer questions about visual quality, text alignment and background consistency.}
\label{fig:survey1}
}
\end{figure}

\section{User Study}
We evaluate our approach with a user study, for 13 text-video pairs and 15 participants. The users were shown source video, and 3-4 methods, randomized (with InVi included), and are expected to answer 3 questions. There are two types of videos: (a) videos for Object swapping and (b) videos for object insertion. In object swapping video, the source video also has the objects, which are modified with a prompt. In Object insertion, the source video do not have the object, and using conditioned control images, we insert a new object in the scene. Moreover, for object swapping videos, we use existing video editing methods are baselines: FateZero~\cite{fatezero}, Tune-A-Video~\cite{wu2023tune} and TokenFlow~\cite{tokenflow2023}. For object insertion, there are no video inpainting pipelines using text-to-image pre-trained models. Hence we use baselines Framewise inpainting (Frm+Inp) and TokenFlow with Controlnet and inpainting pipeline (Tokenflow+Inp). 
For object swapping, we ask users the following questions:
\begin{itemize}
    \item Which video demonstrates the highest consistency with respect to the source video?
\\ \textcolor{blue}{Choose the method which is BEST at preserving the background from the source video.}
\item Which video aligns most accurately with the provided text prompt?
\\ \textcolor{blue}{Choose the method which BEST captures the details in the prompt (given on top of the video).}
\item Which video demonstrates the highest visual quality? 
\\ \textcolor{blue}{Choose a video which has the LEAST amount of extra artifacts (jitter, unwanted blobs), blurriness, unrealistic lighting and flickering.}
\end{itemize}

\begin{figure}[tbh!]
\centering{
\includegraphics[width=0.85\linewidth]{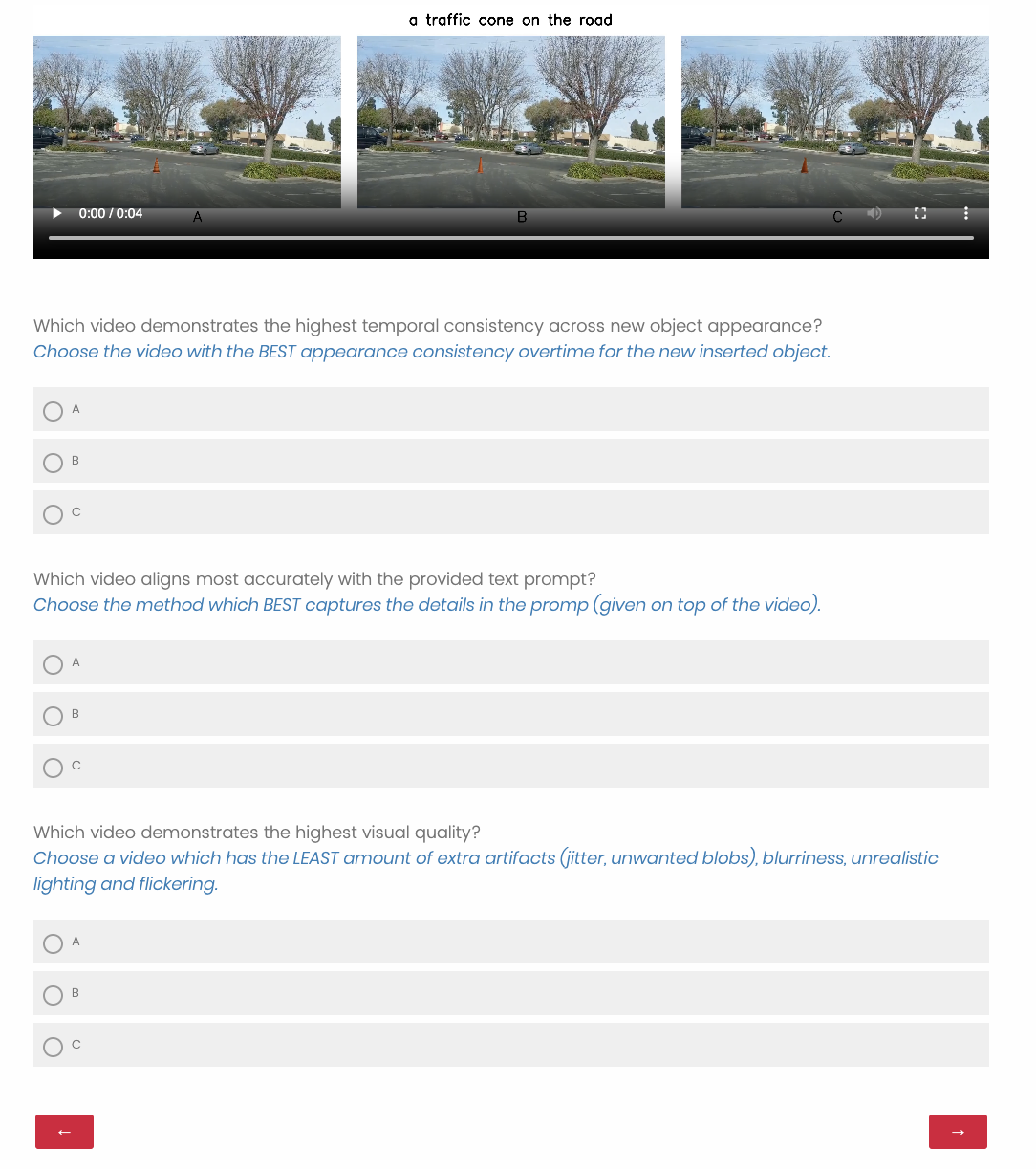}
\caption{Survey Preview for Object Insertion videos. The users are shown 3 videos along with source video and prompt used for editing, to answer questions about overall visual quality, text alignment and temporal consistency.}
\label{fig:survey1}}
\end{figure}
For object insertion, we ask users the following questions:
\begin{itemize}
    \item Which video demonstrates the highest temporal consistency across new object appearance?
\\ \textcolor{blue}{Choose the video with the BEST appearance consistency overtime for the new inserted object.}
\item Which video aligns most accurately with the provided text prompt?
\\ \textcolor{blue}{Choose the method which BEST captures the details in the prompt (given on top of the video).} 
\item Which video demonstrates the highest visual quality? 
\\ \textcolor{blue}{Choose a video which has the LEAST amount of extra artifacts (jitter, unwanted blobs), blurriness, unrealistic lighting and flickering.}
\end{itemize}

%% file: prelims.tex
\section{Preliminaries}
\label{sec:pre}
 We introduce concepts that are required to understand our methods. Readers are encouraged to refer to the methods for a more in-depth treatment.

\noindent \textbf{Diffusion models}~\cite{ho2020denoising} gradually introduce Gaussian noise to a sample $\mathbf{x}_0 \sim q(\mathbf{x}_0)$ over $T$ steps, yielding noisy samples $\mathbf{x}_t, t = 1, \dots, T$. The distribution of these noisy samples is governed by
$ q(\mathbf x_t|\vx_{t-1}) = \mathcal{N}(\mathbf x_t;\sqrt{\alpha_t}\vx_{t-1},\beta_t \mathbf{I})
$, 
where $\beta_t$ denotes the noise variance at a diffusion step $t$ and $\alpha_t = 1 - \beta_t$. Eventually, this \emph{forward process} leads to $\mathbf{x}_T \sim \mathcal{N}(\mathbf{0},\mathbf{I})$, rendering the image $\mathbf{x}_T$ as white noise. Conversely, the \emph{reverse process} inversely applies the aforementioned procedure through the $\theta$-parameterized Gaussian distribution:
$    p_\theta(\vx_{t-1}|\vx_t) = \mathcal{N}(\vx_{t-1};\mu_\theta(\vx_t,t),\beta_t \mathbf{I}).
$
The learning involves estimating $\mu_\theta$ to be able to generate a data sample from noise in $T$ reverse process steps.

\noindent \textbf{Latent Diffusion Models (LDM)}~\cite{rombach2022high} improved the learning and generation process by shifting it from the image space to a latent space. 
The image is encoded into the latent space using an encoder, $E$, and both the forward and reverse diffusion processes occur in this latent space.
The latent space samples are converted back into an image samples using a decoder, $D$. 
The denoising model, based on the U-Net architecture~\cite{ronneberger2015u}, is composed of  self-attention layers~\cite{vaswani} and cross-attention layers~\cite{vaswani} to seamlessly integrate textual conditions. 
These models are also referred to as text-to-image models as a text prompt can be converted into tokens and used within cross attention layers of the U-Net model.

\noindent \textbf{DreamBooth-LoRA based fine-tuning}~\cite{ruiz2023dreambooth, hu2022lora} helps personalize the diffusion model by creating a unique prompt and ``binding'' it with a specific image.
To achieve this ``binding'', first, we generate a prompt with a unique identifier: ``a [V] [class noun]'', where [V] denotes a unique identifier linked to the subject and [class noun] represents a coarse class descriptor of the subject (e.g. boy, horse, etc.).
Next, we condition the diffusion model on this prompt and fine-tune it using the LoRA~\cite{hu2022lora} technique, ensuring that the prompt aligns with the provided image. 
LoRA involves creating a duplicate set of the original diffusion weights, representing them with low-rank matrices, and exclusively training these low-rank matrices while maintaining the original network's frozen state.
The training low-rank matrices are then merged with the original frozen weights, preserving the architecture and keeping the inference time identical to the original model. This approach is the same for inpainting based diffusion models.

\noindent \textbf{DDIM Inversion} (DDIM-Inv) converts a clean sample $\mathbf{x}_0$ to its 
noisy version in reverse steps:
\begin{equation*}
\small
\mathbf{z}_{t+1} = \sqrt{\alpha_{t+1}}\frac{\mathbf{z}_t-\sqrt{1-\alpha_{t}}\epsilon_\theta(\mathbf{z}_t,t,p)}{\sqrt{\alpha_{t}}} + \sqrt{1-\alpha_{t+1}}\epsilon_\theta(\mathbf{z}_{t},t,p), \quad t =0,\dots,T-1.
\end{equation*}
The difference between the forward diffusion process (FDP) and DDIM-Inv is in the noise generation mechanism. 
In the FDP, noise is sampled from a Gaussian distribution, whereas in DDIM-Inv, the noise is the output of the U-Net model.

%% file: main.bbl
\begin{thebibliography}{36}
\providecommand{\natexlab}[1]{#1}
\providecommand{\url}[1]{\texttt{#1}}
\expandafter\ifx\csname urlstyle\endcsname\relax
  \providecommand{\doi}[1]{doi: #1}\else
  \providecommand{\doi}{doi: \begingroup \urlstyle{rm}\Url}\fi

\bibitem[Brooks et~al.(2023)Brooks, Holynski, and Efros]{brooks2023instructpix2pix}
Tim Brooks, Aleksander Holynski, and Alexei~A Efros.
\newblock Instructpix2pix: Learning to follow image editing instructions.
\newblock In \emph{Proceedings of the IEEE/CVF Conference on Computer Vision and Pattern Recognition}, pages 18392--18402, 2023.

\bibitem[Cao et~al.(2023)Cao, Wang, Qi, Shan, Qie, and Zheng]{cao2023masactrl}
Mingdeng Cao, Xintao Wang, Zhongang Qi, Ying Shan, Xiaohu Qie, and Yinqiang Zheng.
\newblock Masactrl: Tuning-free mutual self-attention control for consistent image synthesis and editing.
\newblock \emph{arXiv preprint arXiv:2304.08465}, 2023.

\bibitem[Chen et~al.(2023)Chen, Wu, Xie, Wu, Li, Xia, Xiao, and Lin]{controlavideo}
Weifeng Chen, Jie Wu, Pan Xie, Hefeng Wu, Jiashi Li, Xin Xia, Xuefeng Xiao, and Liang-Jin Lin.
\newblock Control-a-video: Controllable text-to-video generation with diffusion models.
\newblock \emph{ArXiv}, abs/2305.13840, 2023.
\newblock URL \url{https://api.semanticscholar.org/CorpusID:258841645}.

\bibitem[Esser et~al.(2023)Esser, Chiu, Atighehchian, Granskog, and Germanidis]{esser2023structure}
Patrick Esser, Johnathan Chiu, Parmida Atighehchian, Jonathan Granskog, and Anastasis Germanidis.
\newblock Structure and content-guided video synthesis with diffusion models.
\newblock In \emph{Proceedings of the IEEE/CVF International Conference on Computer Vision}, pages 7346--7356, 2023.

\bibitem[Geyer et~al.(2023)Geyer, Bar-Tal, Bagon, and Dekel]{tokenflow2023}
Michal Geyer, Omer Bar-Tal, Shai Bagon, and Tali Dekel.
\newblock Tokenflow: Consistent diffusion features for consistent video editing.
\newblock \emph{arXiv preprint arxiv:2307.10373}, 2023.

\bibitem[Gu et~al.(2023)Gu, Zhou, Wu, Yu, Liu, Zhao, Wu, Zhang, Shou, and Tang]{vid_swap}
Yuchao Gu, Yipin Zhou, Bichen Wu, Licheng Yu, Jia-Wei Liu, Rui Zhao, Jay~Zhangjie Wu, David~Junhao Zhang, Mike~Zheng Shou, and Kevin Tang.
\newblock Videoswap: Customized video subject swapping with interactive semantic point correspondence.
\newblock \emph{ArXiv}, abs/2312.02087, 2023.
\newblock URL \url{https://api.semanticscholar.org/CorpusID:265609343}.

\bibitem[Guo et~al.(2023{\natexlab{a}})Guo, Yang, Rao, Wang, Qiao, Lin, and Dai]{animatediff}
Yuwei Guo, Ceyuan Yang, Anyi Rao, Yaohui Wang, Y.~Qiao, Dahua Lin, and Bo~Dai.
\newblock Animatediff: Animate your personalized text-to-image diffusion models without specific tuning.
\newblock \emph{ArXiv}, abs/2307.04725, 2023{\natexlab{a}}.
\newblock URL \url{https://api.semanticscholar.org/CorpusID:259501509}.

\bibitem[Guo et~al.(2023{\natexlab{b}})Guo, Yang, Rao, Wang, Qiao, Lin, and Dai]{guo2023animatediff}
Yuwei Guo, Ceyuan Yang, Anyi Rao, Yaohui Wang, Yu~Qiao, Dahua Lin, and Bo~Dai.
\newblock Animatediff: Animate your personalized text-to-image diffusion models without specific tuning.
\newblock \emph{arXiv preprint arXiv:2307.04725}, 2023{\natexlab{b}}.

\bibitem[Hertz et~al.(2022)Hertz, Mokady, Tenenbaum, Aberman, Pritch, and Cohen-Or]{hertz2022prompt}
Amir Hertz, Ron Mokady, Jay Tenenbaum, Kfir Aberman, Yael Pritch, and Daniel Cohen-Or.
\newblock Prompt-to-prompt image editing with cross attention control.
\newblock \emph{arXiv preprint arXiv:2208.01626}, 2022.

\bibitem[Ho et~al.(2020)Ho, Jain, and Abbeel]{ho2020denoising}
Jonathan Ho, Ajay Jain, and Pieter Abbeel.
\newblock Denoising diffusion probabilistic models.
\newblock \emph{Advances in neural information processing systems}, 33:\penalty0 6840--6851, 2020.

\bibitem[Hu et~al.(2022)Hu, Shen, Wallis, Allen-Zhu, Li, Wang, Wang, and Chen]{hu2022lora}
Edward~J Hu, Yelong Shen, Phillip Wallis, Zeyuan Allen-Zhu, Yuanzhi Li, Shean Wang, Lu~Wang, and Weizhu Chen.
\newblock Lo{RA}: Low-rank adaptation of large language models.
\newblock In \emph{International Conference on Learning Representations}, 2022.

\bibitem[Khachatryan et~al.(2023)Khachatryan, Movsisyan, Tadevosyan, Henschel, Wang, Navasardyan, and Shi]{khachatryan2023text2video}
Levon Khachatryan, Andranik Movsisyan, Vahram Tadevosyan, Roberto Henschel, Zhangyang Wang, Shant Navasardyan, and Humphrey Shi.
\newblock Text2video-zero: Text-to-image diffusion models are zero-shot video generators.
\newblock \emph{arXiv preprint arXiv:2303.13439}, 2023.

\bibitem[Kirillov et~al.(2023)Kirillov, Mintun, Ravi, Mao, Rolland, Gustafson, Xiao, Whitehead, Berg, Lo, et~al.]{kirillov2023segment}
Alexander Kirillov, Eric Mintun, Nikhila Ravi, Hanzi Mao, Chloe Rolland, Laura Gustafson, Tete Xiao, Spencer Whitehead, Alexander~C Berg, Wan-Yen Lo, et~al.
\newblock Segment anything.
\newblock \emph{arXiv preprint arXiv:2304.02643}, 2023.

\bibitem[Liu et~al.(2023{\natexlab{a}})Liu, Zhang, Li, Lin, and Jia]{liu2023video}
Shaoteng Liu, Yuechen Zhang, Wenbo Li, Zhe Lin, and Jiaya Jia.
\newblock Video-p2p: Video editing with cross-attention control.
\newblock \emph{arXiv preprint arXiv:2303.04761}, 2023{\natexlab{a}}.

\bibitem[Liu et~al.(2023{\natexlab{b}})Liu, Zeng, Ren, Li, Zhang, Yang, Li, Yang, Su, Zhu, et~al.]{liu2023grounding}
Shilong Liu, Zhaoyang Zeng, Tianhe Ren, Feng Li, Hao Zhang, Jie Yang, Chunyuan Li, Jianwei Yang, Hang Su, Jun Zhu, et~al.
\newblock Grounding dino: Marrying dino with grounded pre-training for open-set object detection.
\newblock \emph{arXiv preprint arXiv:2303.05499}, 2023{\natexlab{b}}.

\bibitem[Lugmayr et~al.(2022)Lugmayr, Danelljan, Romero, Yu, Timofte, and Van~Gool]{lugmayr2022repaint}
Andreas Lugmayr, Martin Danelljan, Andres Romero, Fisher Yu, Radu Timofte, and Luc Van~Gool.
\newblock Repaint: Inpainting using denoising diffusion probabilistic models.
\newblock In \emph{Proceedings of the IEEE/CVF Conference on Computer Vision and Pattern Recognition}, pages 11461--11471, 2022.

\bibitem[Oh et~al.(2011)Oh, Hoogs, Perera, Cuntoor, Chen, Lee, Mukherjee, Aggarwal, Lee, Davis, Swears, Wang, Ji, Reddy, Shah, Vondrick, Pirsiavash, Ramanan, Yuen, Torralba, Song, Fong, Roy-Chowdhury, and Desai]{VIRAT}
Sangmin Oh, Anthony~J. Hoogs, A.~G.~Amitha Perera, Naresh~P. Cuntoor, Chia-Chih Chen, Jong~Taek Lee, Saurajit Mukherjee, Jake~K. Aggarwal, Hyungtae Lee, Larry~S. Davis, Eran Swears, Xiaoyang Wang, Qiang Ji, Kishore~K. Reddy, Mubarak Shah, Carl Vondrick, Hamed Pirsiavash, Deva Ramanan, Jenny Yuen, Antonio Torralba, Bi~Song, Anesco Fong, Amit~K. Roy-Chowdhury, and Mita Desai.
\newblock A large-scale benchmark dataset for event recognition in surveillance video.
\newblock \emph{CVPR 2011}, pages 3153--3160, 2011.
\newblock URL \url{https://api.semanticscholar.org/CorpusID:263882069}.

\bibitem[Perazzi et~al.(2016)Perazzi, Pont-Tuset, McWilliams, Gool, Gross, and Sorkine-Hornung]{davis}
Federico Perazzi, Jordi Pont-Tuset, Brian McWilliams, Luc~Van Gool, Markus~H. Gross, and Alexander Sorkine-Hornung.
\newblock A benchmark dataset and evaluation methodology for video object segmentation.
\newblock \emph{2016 IEEE Conference on Computer Vision and Pattern Recognition (CVPR)}, pages 724--732, 2016.
\newblock URL \url{https://api.semanticscholar.org/CorpusID:1949934}.

\bibitem[PNVR et~al.(2023)PNVR, Singh, Ghosh, Siddiquie, and Jacobs]{PNVR_2023_ICCV}
Koutilya PNVR, Bharat Singh, Pallabi Ghosh, Behjat Siddiquie, and David Jacobs.
\newblock Ld-znet: A latent diffusion approach for text-based image segmentation.
\newblock In \emph{Proceedings of the IEEE/CVF International Conference on Computer Vision (ICCV)}, pages 4157--4168, October 2023.

\bibitem[Qi et~al.(2023)Qi, Cun, Zhang, Lei, Wang, Shan, and Chen]{fatezero}
Chenyang Qi, Xiaodong Cun, Yong Zhang, Chenyang Lei, Xintao Wang, Ying Shan, and Qifeng Chen.
\newblock Fatezero: Fusing attentions for zero-shot text-based video editing.
\newblock \emph{2023 IEEE/CVF International Conference on Computer Vision (ICCV)}, pages 15886--15896, 2023.
\newblock URL \url{https://api.semanticscholar.org/CorpusID:257557738}.

\bibitem[Ramesh et~al.(2022)Ramesh, Dhariwal, Nichol, Chu, and Chen]{ramesh2022hierarchical}
Aditya Ramesh, Prafulla Dhariwal, Alex Nichol, Casey Chu, and Mark Chen.
\newblock Hierarchical text-conditional image generation with clip latents.
\newblock \emph{ArXiv}, abs/2204.06125, 2022.
\newblock URL \url{https://api.semanticscholar.org/CorpusID:248097655}.

\bibitem[Rombach et~al.(2022)Rombach, Blattmann, Lorenz, Esser, and Ommer]{rombach2022high}
Robin Rombach, Andreas Blattmann, Dominik Lorenz, Patrick Esser, and Bj{\"o}rn Ommer.
\newblock High-resolution image synthesis with latent diffusion models.
\newblock In \emph{Proceedings of the IEEE/CVF conference on computer vision and pattern recognition}, pages 10684--10695, 2022.

\bibitem[Ronneberger et~al.(2015)Ronneberger, Fischer, and Brox]{ronneberger2015u}
Olaf Ronneberger, Philipp Fischer, and Thomas Brox.
\newblock U-net: Convolutional networks for biomedical image segmentation.
\newblock In \emph{Medical Image Computing and Computer-Assisted Intervention--MICCAI 2015: 18th International Conference, Munich, Germany, October 5-9, 2015, Proceedings, Part III 18}, pages 234--241. Springer, 2015.

\bibitem[Ruiz et~al.(2023)Ruiz, Li, Jampani, Pritch, Rubinstein, and Aberman]{ruiz2023dreambooth}
Nataniel Ruiz, Yuanzhen Li, Varun Jampani, Yael Pritch, Michael Rubinstein, and Kfir Aberman.
\newblock Dreambooth: Fine tuning text-to-image diffusion models for subject-driven generation.
\newblock In \emph{Proceedings of the IEEE/CVF Conference on Computer Vision and Pattern Recognition}, pages 22500--22510, 2023.

\bibitem[Saharia et~al.(2022)Saharia, Chan, Saxena, Li, Whang, Denton, Ghasemipour, Gontijo~Lopes, Karagol~Ayan, Salimans, et~al.]{saharia2022photorealistic}
Chitwan Saharia, William Chan, Saurabh Saxena, Lala Li, Jay Whang, Emily~L Denton, Kamyar Ghasemipour, Raphael Gontijo~Lopes, Burcu Karagol~Ayan, Tim Salimans, et~al.
\newblock Photorealistic text-to-image diffusion models with deep language understanding.
\newblock \emph{Advances in Neural Information Processing Systems}, 35:\penalty0 36479--36494, 2022.

\bibitem[Schuhmann et~al.(2022)Schuhmann, Beaumont, Vencu, Gordon, Wightman, Cherti, Coombes, Katta, Mullis, Wortsman, et~al.]{schuhmann2022laion}
Christoph Schuhmann, Romain Beaumont, Richard Vencu, Cade Gordon, Ross Wightman, Mehdi Cherti, Theo Coombes, Aarush Katta, Clayton Mullis, Mitchell Wortsman, et~al.
\newblock Laion-5b: An open large-scale dataset for training next generation image-text models.
\newblock \emph{Advances in Neural Information Processing Systems}, 35:\penalty0 25278--25294, 2022.

\bibitem[Shrivastava et~al.(2017)Shrivastava, Pfister, Tuzel, Susskind, Wang, and Webb]{Shrivastava_2017_CVPR}
Ashish Shrivastava, Tomas Pfister, Oncel Tuzel, Joshua Susskind, Wenda Wang, and Russell Webb.
\newblock Learning from simulated and unsupervised images through adversarial training.
\newblock In \emph{Proceedings of the IEEE/CVF International Conference on Computer Vision}, 2017.

\bibitem[Suvorov et~al.(2022)Suvorov, Logacheva, Mashikhin, Remizova, Ashukha, Silvestrov, Kong, Goka, Park, and Lempitsky]{suvorov2022resolution}
Roman Suvorov, Elizaveta Logacheva, Anton Mashikhin, Anastasia Remizova, Arsenii Ashukha, Aleksei Silvestrov, Naejin Kong, Harshith Goka, Kiwoong Park, and Victor Lempitsky.
\newblock Resolution-robust large mask inpainting with fourier convolutions.
\newblock In \emph{Proceedings of the IEEE/CVF winter conference on applications of computer vision}, pages 2149--2159, 2022.

\bibitem[Tumanyan et~al.(2022)Tumanyan, Geyer, Bagon, and Dekel]{pnp}
Narek Tumanyan, Michal Geyer, Shai Bagon, and Tali Dekel.
\newblock Plug-and-play diffusion features for text-driven image-to-image translation.
\newblock \emph{2023 IEEE/CVF Conference on Computer Vision and Pattern Recognition (CVPR)}, pages 1921--1930, 2022.
\newblock URL \url{https://api.semanticscholar.org/CorpusID:253801961}.

\bibitem[Tumanyan et~al.(2023)Tumanyan, Geyer, Bagon, and Dekel]{tumanyan2023plug}
Narek Tumanyan, Michal Geyer, Shai Bagon, and Tali Dekel.
\newblock Plug-and-play diffusion features for text-driven image-to-image translation.
\newblock \emph{2023 IEEE/CVF Conference on Computer Vision and Pattern Recognition (CVPR)}, pages 1921--1930, 2023.
\newblock URL \url{https://api.semanticscholar.org/CorpusID:253801961}.

\bibitem[Vaswani et~al.(2017)Vaswani, Shazeer, Parmar, Uszkoreit, Jones, Gomez, Kaiser, and Polosukhin]{vaswani}
Ashish Vaswani, Noam~M. Shazeer, Niki Parmar, Jakob Uszkoreit, Llion Jones, Aidan~N. Gomez, Lukasz Kaiser, and Illia Polosukhin.
\newblock Attention is all you need.
\newblock In \emph{Neural Information Processing Systems}, 2017.
\newblock URL \url{https://api.semanticscholar.org/CorpusID:13756489}.

\bibitem[Wang et~al.(2023)Wang, Yuan, Zhang, Chen, Wang, Zhang, Shen, Zhao, and Zhou]{wang2023videocomposer}
Xiang Wang, Hangjie Yuan, Shiwei Zhang, Dayou Chen, Jiuniu Wang, Yingya Zhang, Yujun Shen, Deli Zhao, and Jingren Zhou.
\newblock Videocomposer: Compositional video synthesis with motion controllability.
\newblock \emph{arXiv preprint arXiv:2306.02018}, 2023.

\bibitem[Wu et~al.(2023)Wu, Ge, Wang, Lei, Gu, Shi, Hsu, Shan, Qie, and Shou]{wu2023tune}
Jay~Zhangjie Wu, Yixiao Ge, Xintao Wang, Stan~Weixian Lei, Yuchao Gu, Yufei Shi, Wynne Hsu, Ying Shan, Xiaohu Qie, and Mike~Zheng Shou.
\newblock Tune-a-video: One-shot tuning of image diffusion models for text-to-video generation.
\newblock In \emph{Proceedings of the IEEE/CVF International Conference on Computer Vision}, pages 7623--7633, 2023.

\bibitem[Yang et~al.(2023)Yang, Zhou, Liu, and Loy]{yang2023rerender}
Shuai Yang, Yifan Zhou, Ziwei Liu, and Chen~Change Loy.
\newblock Rerender a video: Zero-shot text-guided video-to-video translation.
\newblock \emph{arXiv preprint arXiv:2306.07954}, 2023.

\bibitem[Zhang et~al.(2023)Zhang, Rao, and Agrawala]{controlnet}
Lvmin Zhang, Anyi Rao, and Maneesh Agrawala.
\newblock Adding conditional control to text-to-image diffusion models.
\newblock \emph{2023 IEEE/CVF International Conference on Computer Vision (ICCV)}, pages 3813--3824, 2023.
\newblock URL \url{https://api.semanticscholar.org/CorpusID:256827727}.

\bibitem[Zhang et~al.(2018)Zhang, Isola, Efros, Shechtman, and Wang]{lpips}
Richard Zhang, Phillip Isola, Alexei~A. Efros, Eli Shechtman, and Oliver Wang.
\newblock The unreasonable effectiveness of deep features as a perceptual metric.
\newblock In \emph{Proceedings of the IEEE Conference on Computer Vision and Pattern Recognition (CVPR)}, June 2018.

\end{thebibliography}
